\documentclass[runningheads]{llncs}
\usepackage[utf8]{inputenc}
\usepackage[square,sort,comma,numbers]{natbib}
\usepackage{amssymb}
\usepackage{amsmath}
\usepackage{booktabs}
\usepackage{graphicx}
\usepackage{multicol}
\usepackage{multirow}
\usepackage{graphics}
\usepackage{subcaption}
\usepackage{hyperref}

\title{Evaluating  Local Model-Agnostic Explanations of Learning to Rank Models with Decision Paths}
\titlerunning{Evaluating  Local Explanations of LTR models}

\author{Amir Hossein Akhavan Rahnama \inst{1} \and
Judith B\"utepage\inst{1}}
\authorrunning{Rahnama and B\"utepage}
\institute{KTH Royal Institute of Technology}
\begin{document}

\maketitle

\begin{abstract}
Local explanations of learning--to--rank (LTR) models are thought to extract the most important features that contribute to the ranking  predicted by the LTR model for a single data point. Evaluating the accuracy of such explanations is challenging since the ground truth feature importance scores are not available for most modern LTR models. In this work, we propose a systematic evaluation technique for explanations of LTR models. Instead of using black-box models, such as neural networks, we propose to focus on tree-based LTR models, from which we can extract the ground truth feature importance scores using decision paths.  Once extracted, we can directly compare the ground truth feature importance scores to the feature importance scores generated with explanation techniques. We compare two recently proposed explanation techniques for LTR models when using decision trees and gradient boosting models on the MQ2008 dataset. We show that the explanation accuracy in these techniques can largely vary depending on the explained model and even which data point is explained.

\end{abstract}

\section{Introduction}
Learning-to-rank models are omnipresent in the daily life of web users: search engines, e-commerce, recommendation systems and lately machine translation. Already in 2014, Google received over 63000 search queries per second on any given day and more than 197 million people around the world used Amazon every month \citep{chen2014big}. As the volume and complexity of data is growing, the evolution of learning to ranking models is seeing a positive trend in complexity and size ~\citep{fuhr1989optimum, chen2016xgboost,friedman2003multiple, burges2005learning, koppel2019pairwise}. For both engineers and end-users, it is of importance to be able to understand the logic behind the predictions of these learning-to-rank models, especially as their complexity and influence grows.

Explanation techniques aim to provide information about the decision making process of complex machine learning models (sometimes referred to as black-box models) \citep{rudin2019stop}. Explanations of the black-box models can be local or global. \textit{Local} explanations are a set of important features that contribute to the predicted score of a single instance. \textit{Global} explanations are a set of important features that contribute to the prediction of an entire dataset. In order to evaluate the quality of local explanations, human or functionally-grounded methods are often used \citep{doshi2017towards}. In human evaluation methods, human subjects measure the quality of explanations based on different metrics such as informativeness \citep{doshi2017towards}. Human evaluation methods are essential but costly and slow, especially in the earlier phases of the development of explanation techniques. Functionally-grounded (systematic) evaluation methods solve this problem by making use of proxy measures for measuring the quality of explanations. Since functionally-grounded evaluation methods are scalable, they can be used to improve the quality of these explanation techniques in the early phases of their development and they can function as a filter to exclude low quality explanation techniques from entering the human evaluation process.

The real challenge in functionally-grounded evaluation of local explanations is that there is no directly available ground truth for feature importance scores. Because of this, other proxy measures have been proposed in the literature. In \citep{verma2019lirme}, the authors propose two measures for evaluating the quality of explanations for text-based LTR models: explanation consistency and correctness. Explanation consistency is the robustness of an explanation technique against the change of its hyper-parameters, e.g. the explanation length and sample size. When significant changes in hyper-parameters do not lead to a drastic change in the explanation, explanation consistency is assumed to be high. Explanation correctness, as defined by \citep{verma2019lirme}, measures how many of the features that are included in an explanation are frequently occurring terms in the relevant documents. When the explanation assigns high importance scores to terms that also have high term frequency in the relevant documents, explanation correctness is assumed to be high. The authors motivate the definition of explanation consistency with intuition and no objective arguments about this definition is presented.

We argue that, although the proposed definition of explanation consistency in \citep{verma2019lirme} is plausible, their definition of explanation correctness is unjustifiable. We argue that there are no objective ways to measure whether a learning-to-rank model has learned the raw term frequencies from the data and if so, by how large of a magnitude. In addition, term frequency is rarely used in the training of learning-to-rank models. Since raw term frequency values are not very informative in isolation, higher-level computed representations of these term frequencies in form of BM25 \cite{robertson2009probabilistic} and LMIR \citep{wu2014improving} have traditionally been more widely used in the training process of learning-to-rank models \citep{qin2010letor}. In this work, we propose a measure of explanation correctness\footnote{The notions of correctness and accuracy is used interchangeably in our study.} that is extracted from the intrinsic structures of tree-based learning-to-rank models. In our proposed approach, we extract these ground truth values from the decision path of tree-based models. Specifically, our extracted ground truth is based on the features that are used for splitting in different nodes across the decision path. Based on this, our notion of explanation accuracy (or correctness) is defined as the similarity between a local explanation and our extracted ground truth feature importance scores. We compare the explanation accuracy of two recent explanation techniques for LTR models, namely Locally Interpretable Ranking Model Explanations (LIRME) \cite{verma2019lirme} and Explainable Search (EXS) \cite{singh2019exs}. In our empirical experiments, we use different similarity metrics such as Spearman's rank correlation, Euclidean similarity and Area Under the Curve (AUC) scores. Our study includes the explanations of two types of tree-based LTR models, namely point-wise decision trees and pair-wise LambdaMART \citep{burges2005learning}, a gradient boosting model for learning-to-rank tasks \cite{friedman2001greedy}.

Our key findings are: 1) EXS explanations provide the highest accuracy when explaining decision tree models whereas LIRME provides more accurate explanations for the Lambda MART model. 2) The accuracy of both explanation techniques are poor when Spearman's rank correlation and AUC scores are used. 3) There is no significant relationship between the length of the instance's decision path and the explanation accuracy. 4) There might be a direct relationship between the complexity of the explained model and the explanation accuracy when AUC and Spearman's rank correlation are used as similarity metrics.

In the next section, we discuss an overview of related work. Our proposed method for the systematic evaluation of explanation accuracy is presented in Section \ref{sec:method}. Section \ref{sec:experiment} includes the empirical results of our method on the MQ2008 dataset \cite{qin2010letor}. Finally, we discuss the main findings of our study in Section \ref{sec:discussion}.

\section{Related Work}
The evaluation of local explanations based on extracting ground truth feature importance scores from the intrinsic structures of machine learning models has been proposed by several works.

The work of \cite{rahnama2021evaluation} proposes a method to extract the ground truth feature importance for classification tasks in tabular datasets. In the study, the ground truth is extracted from the log odds ratio of Logistic Regression and Naive Bayes models. 

The idea of using decision paths for finding feature importance scores from Decision Tree and Random Forest models is studied in \cite{breiman2001random, chen2015xgboost}, who focus on global importance, and \cite{saabas2015treeinterpreter}, who propose a variation of this method that can determine local feature importance. Based on the latter, the change in the impurity value of each split feature from each node along the decision path is considered as that feature's contributions to the prediction of a single instance. In \cite{lundberg2020local}, the authors evaluate the approach proposed by \cite{saabas2015treeinterpreter} and conclude that this method can find accurate feature contributions in synthetic datasets. 

Few studies have focused on the functionally-grounded evaluation of learning-to-rank explanation techniques. In \cite{verma2019lirme}, the authors show that LIRME explanations are consistent with regards to explanation length and sample size. In addition, the authors showed that LIRME explanations achieve no more than 30\% in terms of explanation correctness across different explanation and sample sizes. As discussed before, they define explanation correctness as the similarity of explanations to the tokens with highest term frequencies. There are no evaluation criteria available in the work of \cite{singh2019exs} for EXS explanations and to our knowledge, no study has compared the quality of explanations of LIRME and EXS to this date.
\section{Local explanations for Learning to Rank}
\label{sec:explanations}
In this section, we provide an overview of the explanation techniques evaluated in this study. First, we provide the formal notation used in this study in Section \ref{sec:exp:notation}. In Section \ref{sec:exp:lirme}, LIRME explanations are briefly discussed and Section \ref{sec:exp:exs} includes an overview of EXS explanations. Figure \ref{fig:exp_example} shows an example of a local explanation generated with LIRME and EXS. 

\subsection{Notation}
\label{sec:exp:notation}
Let $q$ be a query and $D= \{d_1, ..., d_m \}$ be the set of associated documents with the query $q$. Let $S_D$ be the list of predicted scores for the documents in $D$ by the black-box model $f$. The documents are then ranked based on their predicted score in a descending order and returned to the user. Local explanation techniques provide explanations for the predicted scores (or the predicted rank) of the $i$-th instance, namely $S_{d_i}$, in terms of importance scores assigned to all features in $d_i$. In order to obtain explanations, a sampling technique $\sigma$ creates transformed versions of the instance $d_i$. We aim to explain this new set  $D_i' = \{ d'_{i, j} | j= 1, ..., T\}$ where $d_i' = \sigma(d_i)$ and $T$ is a hyper-parameter. The samples are weighted by a kernel function $k$ based on their distance to instance $d_i$. Lastly, a surrogate model $g$ is trained on $(D_i', S_{D_i'})$ where the samples are weighted according to the kernel function $k$. The final explanation are the weights of the surrogate model $g$.

\subsection{LIRME}
\label{sec:exp:lirme}
LIRME is a local explanation technique that uses a LASSO linear surrogate model to explain the original black-box model. In this technique, all features values in the dataset into independent quartiles. In order to create an interpretable representation of the explained instance, a random number between one and four is drawn uniformly at random for each feature. The value of the features in the interpretable representation is set to one only when the randomly selected number is equal to the quartile that includes the value of that feature in the explained instance (see \cite{garreau2020looking} for more details). The sampled instances $d'_{i, j}$ ($j=1, ..., T$) are weighted with the following function

\begin{equation}
    k(d'_{i, j}, d_i) = \textrm{exp}(-\frac{dist(d_{i, j}, d_i)^2}{h}) \label{eq:kernel}
\end{equation} 

where $dist$ represents the distance between $d'_{i, j}$ and $d_i$ and $h$ is a hyper-parameter. In practise, Euclidean or Cosine distance metrics are used \cite{ribeiro2016should}. The LIRME surrogate loss is defined as follows:

\begin{equation}
    \mathcal{L}(D_i', S_{D_i'}, k) = \sum_{j=1}^{T} k(d'_{i, j}, d_i) (g(d'_{i, j}) - S_{d'_{i, j}})^2 + \alpha |\Theta| 
\end{equation}


where $\Theta$ are the parameters of the surrogate model g. 

\subsection{EXS}
\label{sec:exp:exs}
EXS is a local explanation technique that uses a Support Vector Regression (SVR) surrogate model to explain the original black-box model. EXS can provide explanations for both predicted scores or ranks of an individual instance $d_i$. EXS follows a sampling technique that selects a set of features uniformly at random and sets their values to the average value of that feature in the dataset\footnote{Authors set the term frequency value of the sampled term to zero. We assume that this translates to replacing a value with its empirical average value in tabular datasets.}. Each generated sample, $d'_{i, j}$ is weighted with a kernel function of the same form as in Equation \ref{eq:kernel}.

where $dist$ represents the distance between $d'_{i, j}$ and $d_i$ and $h$ is a hyper-parameter. Unlike LIRME, EXS does not directly utilize the $S_D$ values in the training of the surrogate model. Instead, these values are first transformed into $y'_j$,

\begin{equation*}
    y'_j \gets \frac{S_{d_{\text{max}}} - S_{d'_{i, j}}}{S_{d_{\text{max}}}} 
\end{equation*}
where $j=1, ..., T$ and $S_{d_{\text{max}}}$ is the instance with the highest predicted score based on the $S_{D'}$ values for the query $q$. The surrogate model $g$ is then regressed on the $(d'_{i, j},y'_j)$ pairs and each instance is weighted with the values of $k(d'_{i, j}, d_i)$.  The explanations are then the weights of the surrogate model $g$.

To summarize, There are three main differences between LIRME and EXS techniques: 1) Each explanation technique trains a different class of surrogate models 2) LIRME explains the predicted score of the black-box whereas EXS explains the normalized differences between the predicted score of the explained document and the top rank document in the same query

\section{Methodology}
\label{sec:method}
\begin{figure}[t]
 \centering
  \includegraphics[scale=0.3]{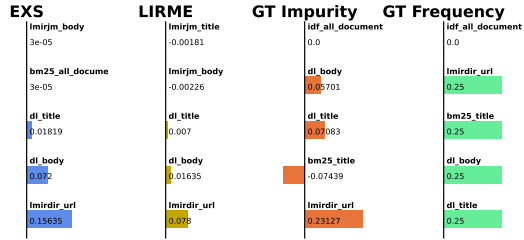}
 \caption{Example of EXS and LIRME explanations for a test instance in MQ2009 dataset. Next to the explanations, the extracted ground truth (GT) from the decision tree model is depicted for that test instance. The visible features are the top-5 ranked features based on their absolute importance scores.}
  \label{fig:exp_example}
 \end{figure}

In this section, we  present our proposed procedure for measuring explanation correctness. Our procedure has two steps: firstly, we extract ground truth feature importance scores from the decision paths that are intrinsically present in tree-based models. After that, we measure the similarity between the extracted ground truth and local explanations. Our methodology include two measures of ground truth that are based on 1) node impurity (see Section \ref{sec:method:impurity}) and 2) feature frequency (see Section \ref{sec:method:frequency}).   In Section \ref{sec:method:similarity}, we discuss the similarity metrics used in our study. Figure \ref{fig:exp_example} shows an example of extracted ground truth for a test instance of MQ2008 based on impurity, i.e. Ground Truth (GT)Impurity, and frequency, i.e. GT Frequency. We demonstrate how to calculate the impurity and frequency based ground truth scores for a point-wise decision tree model in Section \ref{sec:method:example}.

\subsection{Ground Truth: Impurity}
\label{sec:method:impurity}
\citep{saabas2015treeinterpreter} proposed to use the decision path as a way for calculating feature importance scores with respect to the predicted score of a decision tree model for a single instance. The change in the node impurity value from node $A$ to node $B$ along the decision path is assumed to represent the importance of the feature that was used to split node $A$. Leaf nodes are excluded in this process. In case a feature is used by multiple nodes on a decision path, the importance values from all of those occurrences are summed up together. The sum of the bias term\footnote{The bias term in decision tree models is the average value of the response variable for regression tasks and the proportion of instances per each class in the classification tasks.} and all feature importance scores is equal to the predicted score by the decision tree model. \citep{saabas2015treeinterpreter} proposed to average the feature importance scores across all trees for the case of bagging models such as Random Forests. Here, we propose the same approach for extracting importance scores from boosting tree models. The impurity-based ground truth can potentially favor the explanation techniques that directly explain the predicted scores of a black-box model such as LIRME.

\subsection{Ground Truth: Feature Frequency}
\label{sec:method:frequency}
The impurity-based feature importance scores come with biases and limitations. As argued in \citep{lundberg2020local}, this method is biased in favoring features that are only used for splits in nodes with large values of depth. In other words, important features that are used numerous times for splitting across different nodes in a decision paths can receive lower importance scores if they cause smaller changes in node impurity values.  In addition, impurity-based ground truth scores might still be a biased benchmark for evaluating model-agnostic explanation techniques as these techniques do not have any direct access to the internal structure of the tree model they explain, and hence the nodes impurity values. However, we argue that an accurate technique that explains tree-based models needs to deem features along the decision path as important, as no other intrinsic property of a tree-based model except the decision path controls its predicted score.

The idea of using feature frequency for finding feature importance scores in tree models was originally proposed by \citep{chen2015xgboost}, who used it to obtain global explanations. In this study, we propose the use of this measure to evaluate local explanations. Our proposed measure is based on the normalized frequency of features that are used for splits along the nodes in a decision path. We argue that this measures provides a complementary view on important features that contribute to the prediction of a single instance in a decision tree model. Based on our method, the ground truth feature importance scores are averaged across all trees when explaining boosting and bagging trees.We would like to emphasize that unlike the impurity-based ground truth, the sum of all feature's frequency-based ground truth importance scores is not equal to the predicted scores of a black-box in this case. As a result, the frequency-based ground truth can favor techniques that are not directly explaining the predicted scores such as EXS.

\subsection{Similarity Metric}
\label{sec:method:similarity}
There are no generally accepted similarity metrics for measuring explanation accuracy. The measure of choice is dependent on the application use-case and the ways in which explanation are presented to users \citep{montavon2018}. In \citep{ribeiro2016should}, Euclidean and Cosine similarity metrics are used for evaluating the explanation similarity. Moreover, \citep{ghorbani2019interpretation, rahnama2021evaluation} proposed the use of Spearman's rank correlation and \cite{hsieh2020evaluations} argued for the use of AUC scores.

In this study, we make use of three aforementioned similarity metrics, namely AUC scores, Euclidean similarity and Spearman's rank correlation. We follow the proposition of \cite{hsieh2020evaluations} for calculating the AUC scores: we rank the features based on the absolute value of their important scores and create a one-hot vector such that a feature with the rank of less than or equal to $K$ has a value of one. In this case, $K$ is a hyper-parameter that is set by the user. We select $K=5$ as the default value in our study, however we analyze the values of explanation accuracy with various $K$ in Section \ref{sec:experiment}.

\begin{table*}[b!]
\centering
\begin{tabular}{l|c|c|c|c}
     \toprule
    Feature  & \ LMIRJM URL \ & \ LMIR ABS Title \  & \ BM25 Title \ & \ BM25 body \ \\
     \midrule
    Value  &  0.786  & 0.0 & 0.722 & 0.780 \\ 
    \bottomrule
\end{tabular}
\caption{ Example Test Instance from MQ2008 dataset}
\label{table:single_instance_example}
\end{table*}

We argue that Spearman's rank correlation is an optimal metric for evaluating the explanation accuracy in use-cases where the ranked subset of important features are presented to the users, e.g. tabular datasets \citep{molnar2020interpretable, psychoula2021explainable}. Firstly, Spearman's rank correlations is insensitive to small changes in importance scores that do not affect the final rank of features. In addition, this metric can identify cases where the rank of features in an explanation and ground truth are in the opposite directions.

\subsection{Ground Truth: Example}
\label{sec:method:example}
\begin{figure}[t]
 \centering
  \includegraphics[scale=0.25]{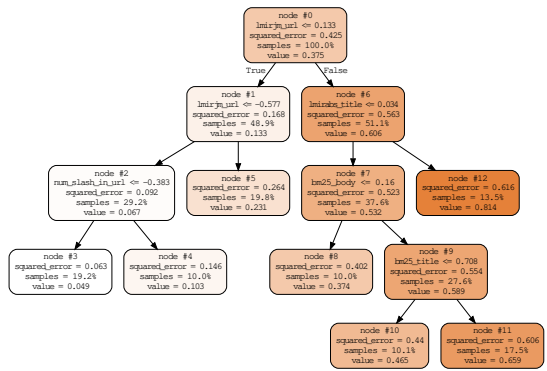}
 \caption{The logic of the decision tree model trained on MQ2008. The darker nodes have larger impurity values (denoted as \textit{values} in each node). The value of sample denotes the percentage of training samples that reach each node.}
  \label{fig:dt_logic}
 \end{figure}

In this section, we show how the ground truth can be extracted for an example instance from the test set of MQ2008 dataset when explaining the decision tree model shown in Figure \ref{fig:dt_logic}. The subset of features that are used to make a decision for this instance are depicted in Table \ref{table:single_instance_example}\footnote{See \cite{qin2010letor} for the list of features and their corresponding descriptions.}. The model used for this example is a point-wise decision tree model. It can be seen that the decision path for the prediction of this individual instance includes root node 0, internal nodes 6, 7, 9 and the leaf node 11. 


\begin{table*}[t!]
\centering
\begin{tabular}{lcc|cc}
    & \multicolumn{2}{c|}{GT Impurity} & \multicolumn{2}{c}{GT Frequency} \\
    \toprule
    Similarity  & LIRME & EXS & LIRME & EXS    \\
    \toprule
     Spearman       &  0.4003 & 0.487  & 0.3898 & 0.296 \\
     Euclidean      &  0.8427 & 0.8937 & 0.6881 & 0.7164 \\
     AUC (Top-10)   &  0.6 & 0.6 & 0.6 & 0.6\\
 \bottomrule \\
\end{tabular}
\caption{Explanation accuracy for a single explained test instance from MQ2008}
\label{table:measure_single_example}
\end{table*}

Let us calculate the impurity-based ground truth feature importance scores of this instance. We can see that the importance of feature LMIRJM URL is 0.231 at the root. Along the decision path, from node 0 to node 6, the node impurity values increases from 0.375 to 0.606 (a positive change) at the beginning of this path. After that, the importance score of feature LMIR ABS Title is -0.074, since the feature causes a decrease in impurity value from 0.606 to 0.532 along the decision path. Following the same logic, the importance of feature BM25 Body is -0.074 (0.532 - 0.606) and lastly, feature BM25 Title has the importance of 0.057 (0.589 - 0.532). The importance of features that do not appear in this path is 0. The calculated ground truth are depicted in Figure \ref{fig:exp_example} as GT Impurity.

After that, we will calculate the frequency-based ground truth scores for this instance. We can see that features LMKIRJM URL, LMIR ABS Title, BM25 TITLE And BM25 BODY are all used once for splitting by nodes along the decision path of the instance (see Figure \ref{fig:dt_logic}). Therefore, they all equally get the importance scores of 0.25 (1/4 = 0.25). The calculated ground truth values are shown in Figure \ref{fig:exp_example} as GT Frequency.

In Table \ref{table:measure_single_example}, we present the result of explanation correctness for this single instance. We can see that EXS outperforms LIRME on the impurity-based ground truth except when the AUC score is used as the similarity metric. In the case of the frequency-based ground truth measure, EXS provides a higher explanation accuracy for Euclidean similarity whereas LIRME outperforms EXS for Spearman's rank correlation. Both techniques provide equal explanation correctness based on top-10 AUC score regardless of the type of ground truth used.

\section{Experiments}
\label{sec:experiment}
In this section, we present our empirical study of comparing explanation accuracy in MQ2008 dataset. The code for experiments is available at: \url{https://anonymous.4open.science/r/ltr_explanation_evaluation}. After detailing the dataset and models in Section \ref{sec:experiment:data} and Section \ref{sec:experiment:models}, we analyze the accuracy of the different techniques and models in Section \ref{sec:experiment:accuracy}. In the following, we analyze the influence of the choice of $K$ on AUC scores in Section \ref{sec:experiment:k}. Finally, we investigate the impact of the length of the decision path on the accuracy in Section \ref{sec:experiment:depth}.

\subsection{Data}
\label{sec:experiment:data}
The MQ2008 dataset \cite{qin2010letor} has 471 training queries with 5938 data instances and 157 validation queries with 1935 data instances. We use all the 156 queries with 1963 data instances in the test set for obtaining and analysing the explanation accuracy. Each document consists of 49 features of which are numerical (detailed description of features are available in \cite{qin2010letor}).

\subsection{Model details}
\label{sec:experiment:models}
We compare the two existing explanation techniques for LTR, LIRME and EXS, on a decision tree model and on a pairwise LambdaMART model. LambdaMart \citep{wu2010adapting} is an adaptive boosting model based on Multiple Additive Regression Trees (MART) \citep{friedman2001greedy} and is a widely used LTR model \citep{chapelle2011yahoo}. Based on our knowledge, this is the first study that focuses on the evaluation of local explanations of the LambdaMART model.

We perform cross-validation with random hyper-parameter optimization for finding the most optimal point-wise decision tree model. The best estimator out of 1000 trials is a decision tree model with a depth of 4 and a test mean squared error of 0.241. The overall logic of tree model is shown in Figure \ref{fig:dt_logic}. We perform similar trials for choosing the most optimal pair-wise LambdaMART model. The best LambdaMART estimator has three trees each of which has a depth of 20. The test NDCG scores of the selected LambdaMART model is 0.066\footnote{More information about the parameters and hyper-parameters of these tree-based models can be found in the appendix.}. The number of samples used for obtaining LIRME and EXS explanations is 2000 as suggested in the original studies.


\subsection{Accuracy comparison}
\label{sec:experiment:accuracy}

We start by comparing the explanation accuracy of LIRME and EXPS explanations for the decision tree model and LambdaMART.

In Table \ref{table:accuracy}, the average and standard deviation in explanation accuracy values of LIRME and EXPS are shown. In the case of the decision tree,  EXS explanations marginally outperform LIRME explanations with respect to all similarity metrics for both impurity and frequency-based ground truth. The explanation accuracy results when Spearman's rank correlation is used can be interpreted as having negligible correlation values according to \cite{schober2018correlation}.  For LambdaMART, we can see that the LIRME explanations outperform EXS explanations with respect to all similarity metrics for both impurity and frequency-based ground truth values. This is in contrast to the results on the decision tree model.

Interestingly, EXS has a higher average accuracy than LIRME under the impurity-based ground truth scores when explaining the decision tree model. This is surprising because EXS, in contrast to LIRME, does not explain the predicted scores. Since the impurity scores are directly linked to the predicted scores, intuitively a model that explains the predicted scores, such as LIRME, should perform better. This is not the case.  

Similarly, LIRME explanations for LambdaMART have slightly higher accuracy under the frequency-based ground truth scores even though LIRME explains the predicted score. One possible conclusion is that the decomposed value, such as predicted probability score, does not play a strong role in the selection of the important features.

The results in Table \ref{table:accuracy} show that the average explanation accuracy of LIRME and EXS is lower for explanations of LambdaMART relative to the explanations of the decision tree when Spearman's rank correlation and AUC scores are used. One possible conclusion is that there might be a direct relationship between the complexity of the explained model and the average explanation accuracy. LIRME and EXS explanation techniques achieve a higher accuracy based on the impurity ground truth when explaining the decision tree model across all similarity metrics. On the other hand, explanation accuracy values are higher in the case of the frequency ground truth for the LambdaMART model, except when Euclidean similarity is chosen.


\begin{table}
\centering
\caption{The average and standard deviation of explanation accuracy when explaining a decision tree and LambdaMART model on MQ2008 dataset}
\label{table:accuracy}
\resizebox{\columnwidth}{!}{%
\begin{tabular}{c|c|c|c|c|c}
&  & \multicolumn{2}{c|}{GT Impurity} & \multicolumn{2}{c}{GT Frequency}\\
\hline
Model & Similarity           & LIRME              & EXS                & LIRME              & EXS                \\
\hline
 \multirow{2}{*}{Decision Tree} & AUC & 0.652$\pm$0.09 & 0.697$\pm$0.08  & 0.65$\pm$0.12 & 0.68$\pm$0.08 \\
 & Spearman  & 0.293$\pm$0.12 & 0.303$\pm$0.14 & 0.291$\pm$0.12 & 0.30$\pm$0.07 \\
 & Euclidean & 0.825$\pm$0.03 & 0.848$\pm$0.06 & 0.61$\pm$0.06  & 0.626$\pm$0.08 \\
 \midrule
 \multirow{2}{*}{LambdaMART} & AUC   & 0.567$\pm$0.09 & 0.576$\pm$0.09 & 0.582$\pm$0.09 & 0.595$\pm$0.09 \\
 & Spearman  & 0.122$\pm$0.16 & 0.12$\pm$0.15  & 0.142$\pm$0.16 & 0.135$\pm$0.15 \\
 & Euclidean & 0.819$\pm$0.14 & 0.743$\pm$0.15 & 0.75$\pm$0.02  & 0.73$\pm$0.09  \\

\hline
\end{tabular}%
}
\end{table}

\subsection{The choice of K}
\label{sec:experiment:k}
\begin{figure}[t]
 \centering
  \includegraphics[scale=0.5]{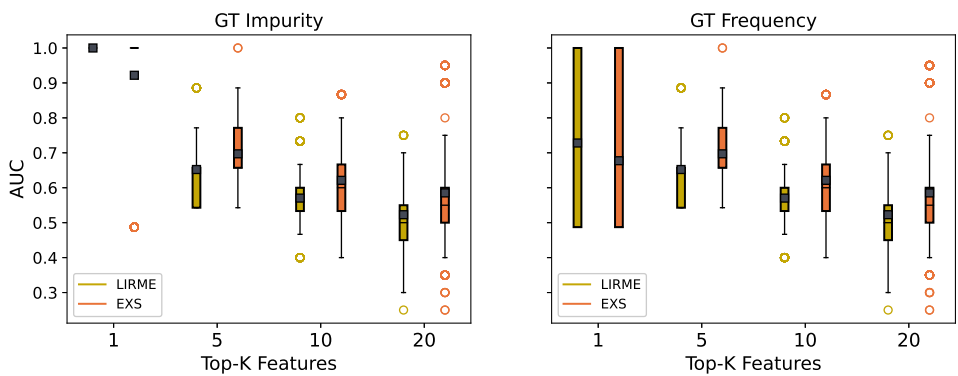}
 \caption{The value of top-$K$ AUC explanation accuracy based on different values of $K$ when explaining the decision tree model. The dark rectangle represents the median in each boxplot.}
  \label{fig:auc_dt}
 \end{figure}

Next, we are interested in the role that the selection of $K$ plays in the estimation of the AUC scores.
In Figure \ref{fig:auc_dt}, we show a boxplot of different values of explanation accuracy for the decision tree model with respect to the top-$K$ AUC scores for various values of $K$. The average AUC values decreases monotonically as the value of $K$ increases while the standard deviation values increase from values 1 to 5 and stay relatively constant for $K$ larger than 10. This trend is similar across both types of the extracted ground truth measures. Interestingly, both explanations techniques can achieve high AUC scores for the top  feature under the impurity-based ground truth.

\begin{figure}[t]
 \centering
  \includegraphics[scale=0.4]{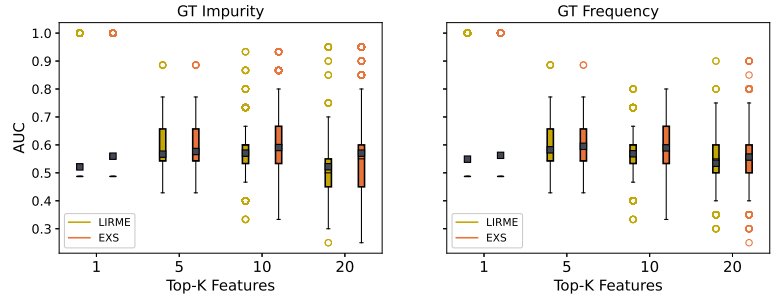}
 \caption{The variation in the explanation accuracy based on top-$K$ AUC scores with different values of $K$ when explaining the LambdaMART model. The dark rectangle represents the median in each boxplot.}
  \label{fig:auc_lmart}
 \end{figure}

Similar behavior can be observed for the LambdaMART model as shown in Figure \ref{fig:auc_lmart}. However, the decreasing trend is not as obvious as in the case of the decision tree.

\subsection{The depth of the decision path}
\label{sec:experiment:depth}
\begin{figure}[!htb]
\minipage{0.32\textwidth}
  \includegraphics[width=\linewidth]{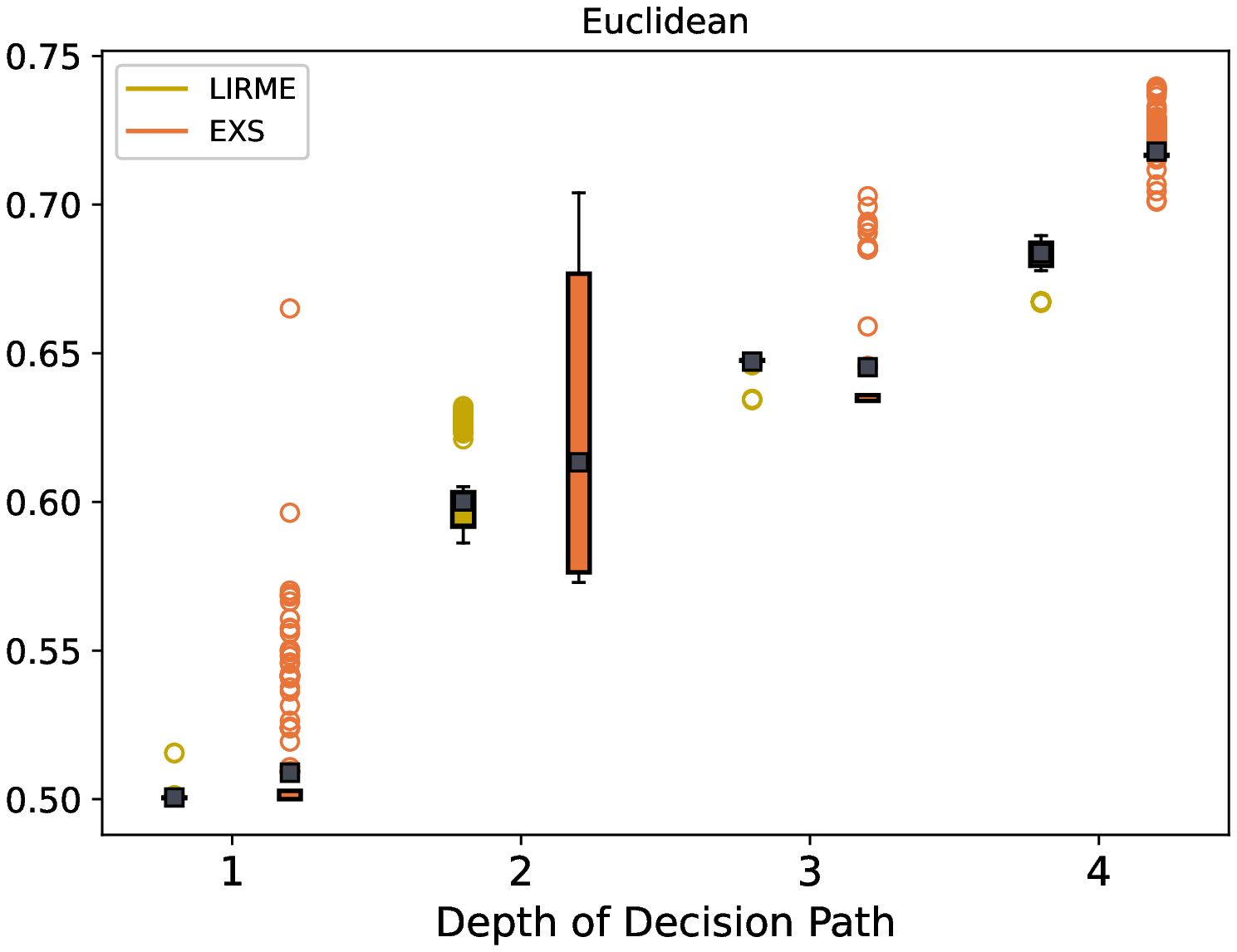}
\endminipage\hfill
\minipage{0.32\textwidth}
  \includegraphics[width=\linewidth]{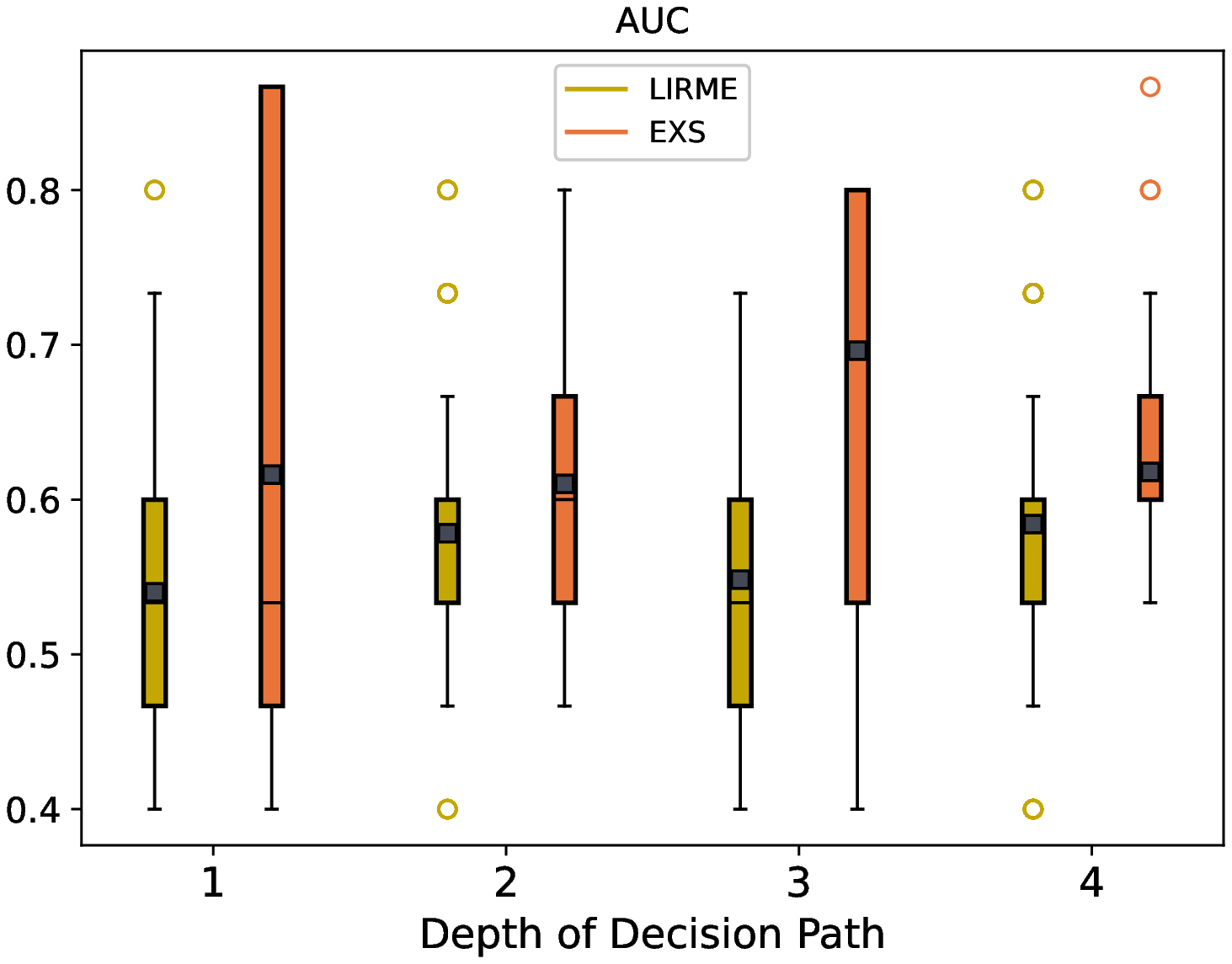}
\endminipage\hfill
\minipage{0.32\textwidth}%
  \includegraphics[width=\linewidth]{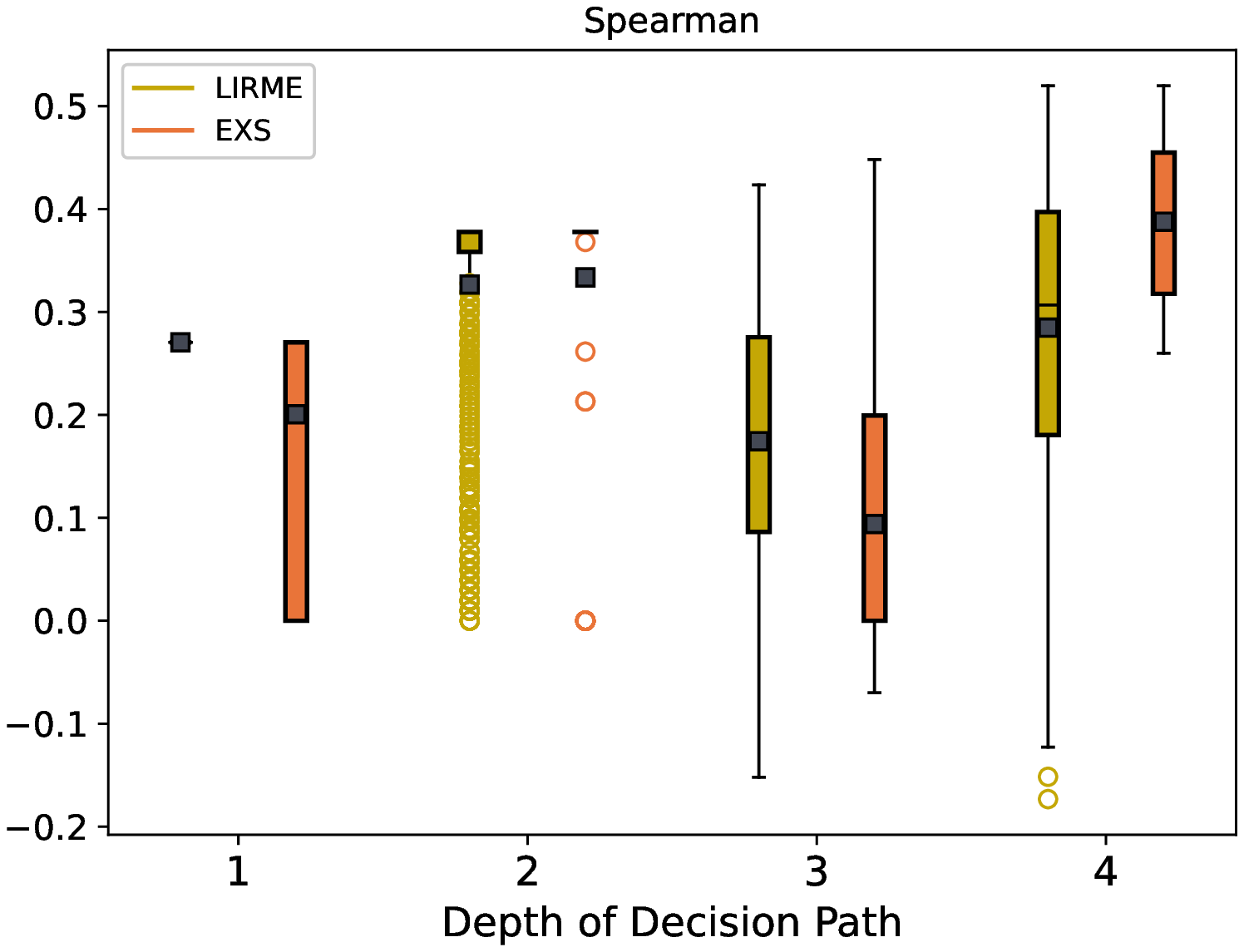}
\endminipage
\caption{Explanation accuracy based on frequency ground truth with respect to the depth of the decision path when explaining the decision tree model. The dark rectangle represents the median in each boxplot.}
 \label{fig:boxplot_path_dt_frequency}
\end{figure}

\begin{figure}[!htb]
\minipage{0.32\textwidth}
  \includegraphics[width=\linewidth]{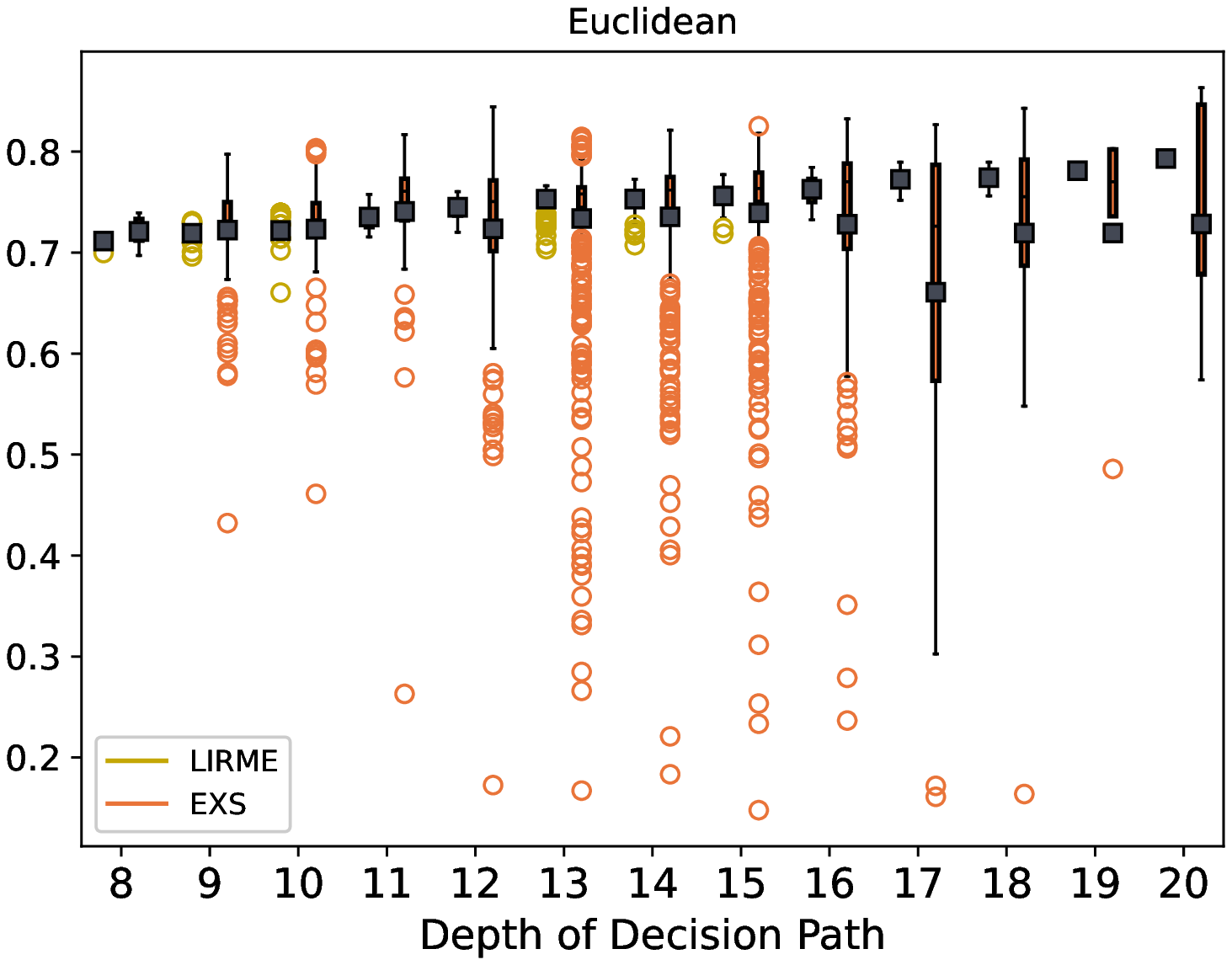}
\endminipage\hfill
\minipage{0.32\textwidth}
  \includegraphics[width=\linewidth]{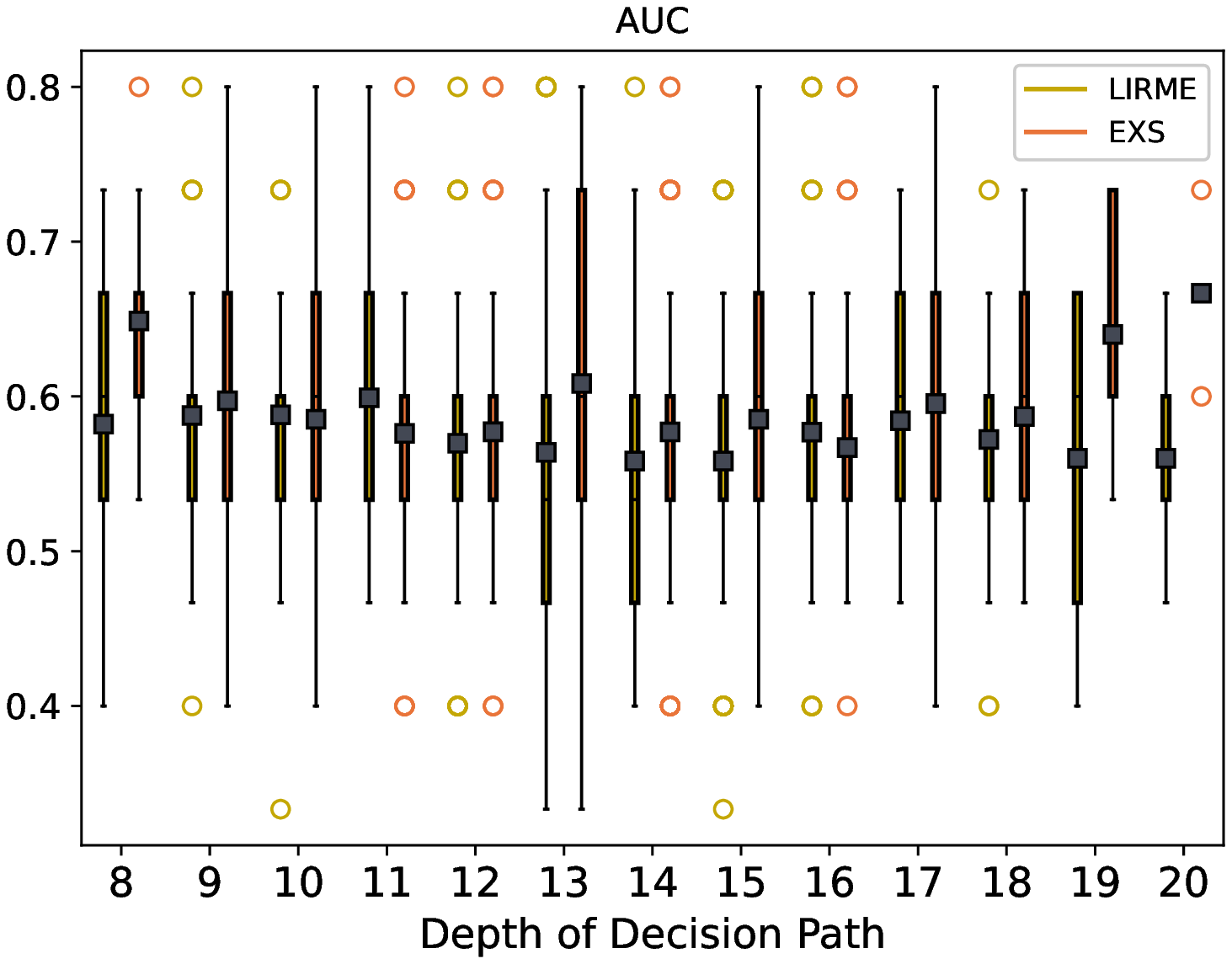}
\endminipage\hfill
\minipage{0.32\textwidth}%
  \includegraphics[width=\linewidth]{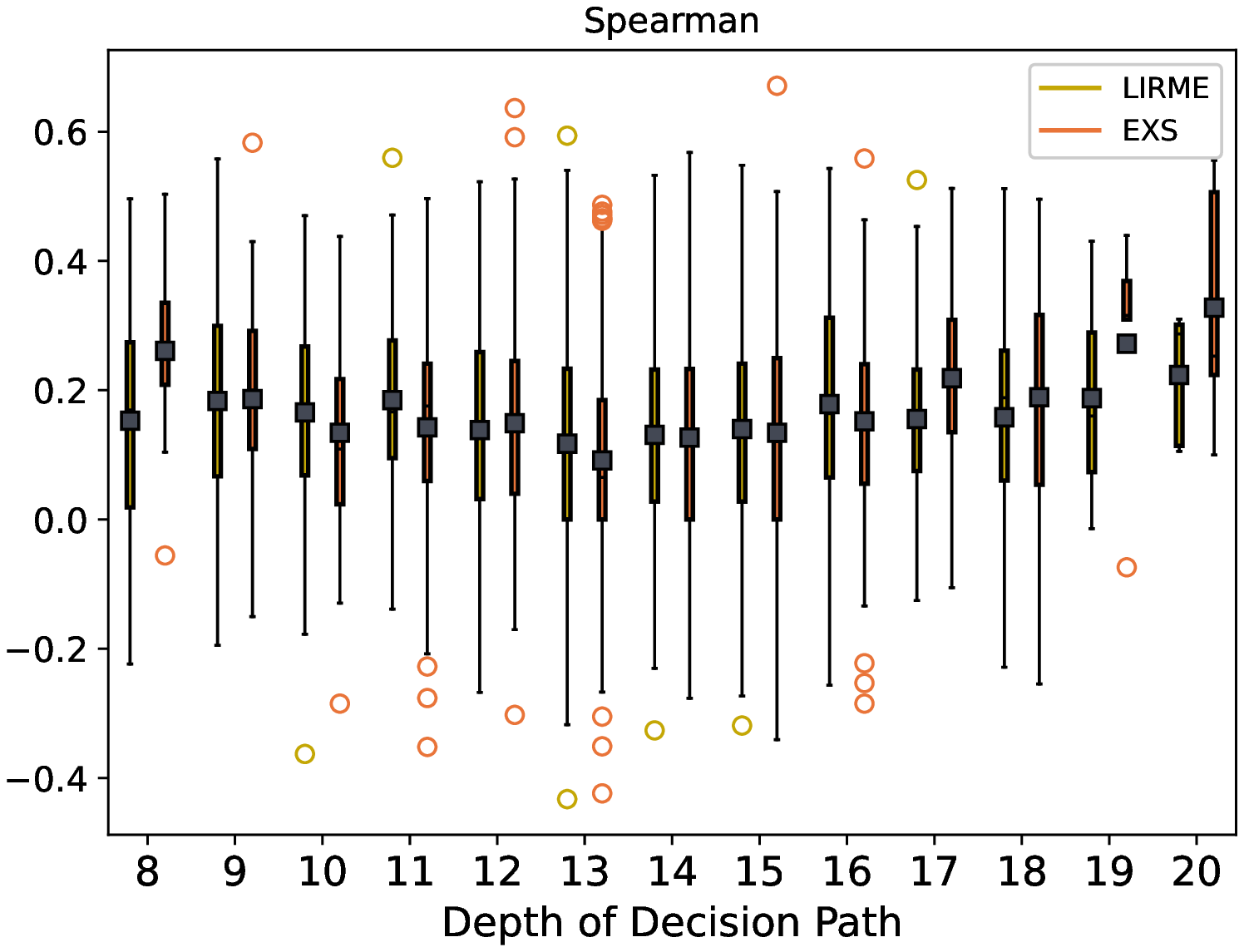}
\endminipage
\caption{Explanation accuracy based on frequency ground truth with respect to the depth of the decision path when explaining the LambdaMART model. The dark rectangle represents the median in each boxplot.}
 \label{fig:boxplot_path_lmart_frequency}
\end{figure}

As previously mentioned, \citep{lundberg2020local} argues that the explanation accuracy values increase with the increase in the depth of the decision path. We investigate this claim by systematically identifying the length of the decision path and correlating it with our accuracy measures.In Figure \ref{fig:boxplot_path_dt_frequency}, the values of explanation accuracy are shown based on the depth of the decision path in the decision tree model for the case of frequency-based ground truth. The test instances with deeper decision paths have more accurate explanations when Euclidean similarity is used. However, no direct relationship between explanation accuracy and the depth of the decision path is visible when AUC scores or Spearman's rank correlation are used.  Similar trends apply to the explanation accuracy of the LambdaMART model based on the frequency ground truth in Figure \ref{fig:boxplot_path_lmart_frequency}. We do not have yet have an explanation for this trend.


Surprisingly, explanation accuracy and the length of the decision path have no direct relationship for either the decision tree or the LambdaMART model when the impurity ground truth is used (see Appendix).

\section{Discussion}
\label{sec:discussion}
Functionally-grounded evaluation of explanation techniques are important since human evaluation methods can be costly. In this work, we proposed a measure of explanation accuracy that is based on extracting ground truth feature importance scores from the decision path in the tree-based models. We base our idea on previous studies that have shown the decision path to be an effective intrinsic property to explain the predictions of a tree-based model, both in terms of global  \citep{chen2015xgboost} or local feature importance scores \citep{saabas2015treeinterpreter, lundberg2020local}. 

Given that LIRME and EXS explanations are agnostic to the internal structures of the models they explain, it can be argued that the decision path is a biased property for evaluating these techniques. To date, the only known intrinsic structure in the tree-based models that encloses the logic of the prediction are decision paths. It is objectively difficult to argue that features which are not used for splitting the nodes in decision tree models can still be important in the prediction process of tree-based models.

Based on our results on the MQ2008 dataset, we observe that the optimal choice of explanation techniques might depend on the underlying black-box model. For example, EXS provides more accurate explanations when explaining the decision tree model. On the other hand, LIRME explanations are more accurate in explaining the Lambda MART model. 

We also found that LIRME and EXS explanations have a relatively higher average accuracy when explaining the decision tree model compared to the Lambda MART model. Based on this, we suspect that explaining more complex models is only possible when the explanation techniques have achieved a relatively high accuracy in explaining simpler models. We also observe that the values of average explanation accuracy based on Spearman's rank correlation are "negligible" when explaining the decision tree and LambdaMART model, e.g. less than 0.3 when explaining the decision tree model and less than 0.15 when LambdaMART model. Based on this, we argue that LIRME and EXS explanations need further improvements in their design for use-cases where the accurate rank of important features is crucial, as is the case with tabular datasets.

We showed that the relationship between the depth of decision path and explanation accuracy, as shown in \cite{lundberg2020local}, are only visible in cases where the frequency-based ground truth and Euclidean similarity are used.

Explanation accuracy values have large standard deviations when using Spearman's correlation across both measures of ground truth scores. In such cases, we can conclude that local explanation techniques cannot explain all single instances correctly. We argue that it is vital for explanation techniques to have intrinsic criteria for when not to explain the prediction of an individual instance. To the best of our knowledge, such criteria are not present in local explanation techniques for classification or learning-to-rank tasks.

In our study, the relative rank of explanation techniques, based on their average explanation accuracy across the entire dataset, was constant across different similarity metrics. However, we would like to emphasize that the choice of similarity metrics is crucial in measuring explanation accuracy. To date, there is no globally optimal similarity metric for measuring explanation accuracy.
 
Our method for measuring explanation correctness can only be used when explaining the local decisions of tree-based models. We believe this to be the main limitation of our proposed approach. Even though tree-based models are nonlinear and numerous models in learning-to-rank challenges are tree-based, neural network models have achieved the state-of-the-art accuracy across numerous learning-to-rank tasks. The neural network state-of-the-art models have shown significant accuracy scores in tasks such as image search \cite{cao2020unifying} and question answering \cite{srivastava2020visual}. Our proposed approach cannot measure explanation correctness in cases when neural network models are explained. Therefore, one possible future direction of this study is to extract ground truth scores from intrinsic structures of neural networks models. The other limitation of our approach is that we cannot measure the explanation accuracy when there is no direct access to the internal structure of the explained tree-based model.  

Lastly, We would like to emphasize the importance of evaluating explanation accuracy with human-in-the-loop techniques. It is important to understand, that although our proposed measure of explanation correctness is systematic and scalable, systematic measures can by no means replace human evaluation studies and are only complementary to them.

\bibliographystyle{splncs04}
\bibliography{references}
\newpage
\section*{Appendix}

\begin{figure}[!htb]
\minipage{0.32\textwidth}
  \includegraphics[width=\linewidth]{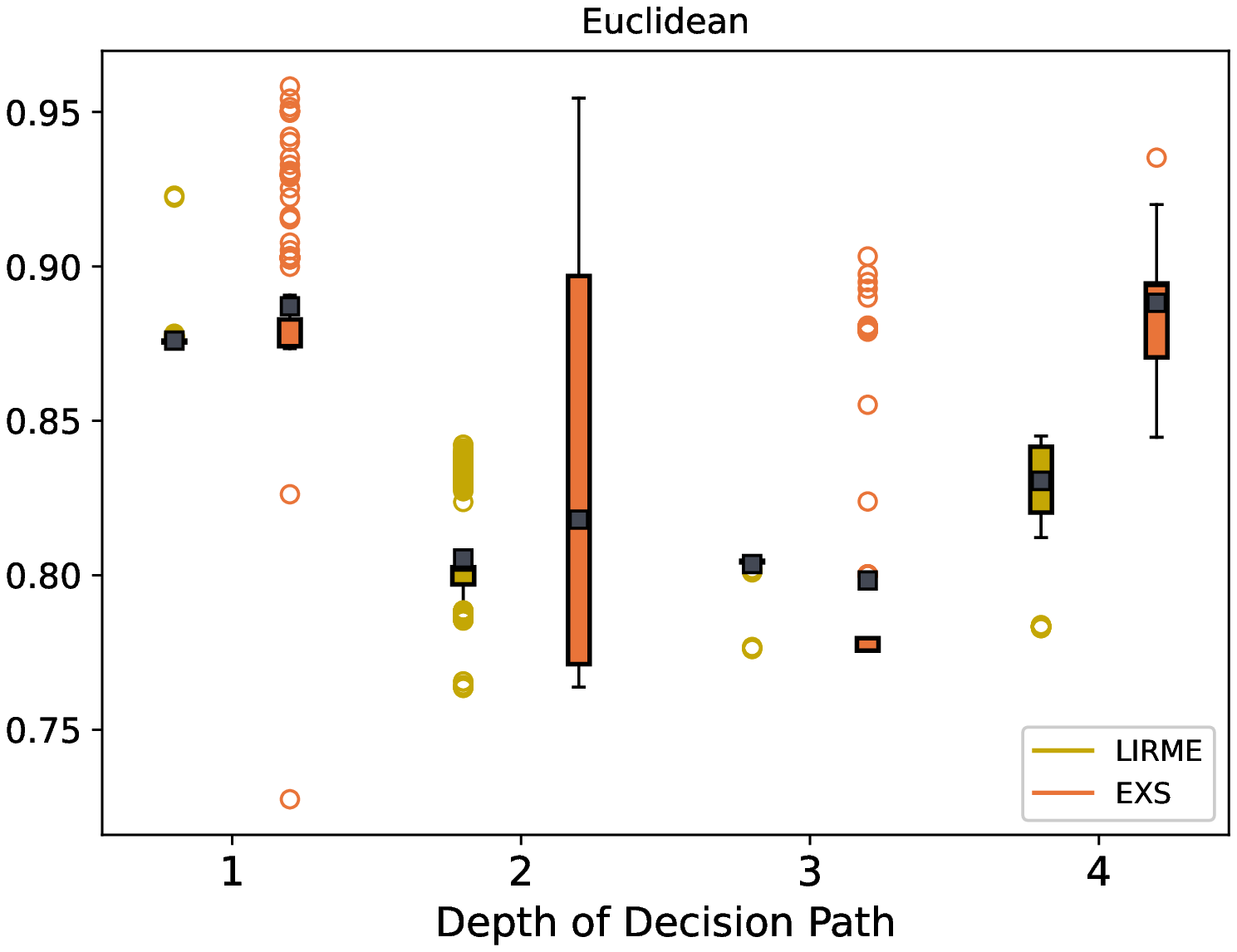}
\endminipage\hfill
\minipage{0.32\textwidth}
  \includegraphics[width=\linewidth]{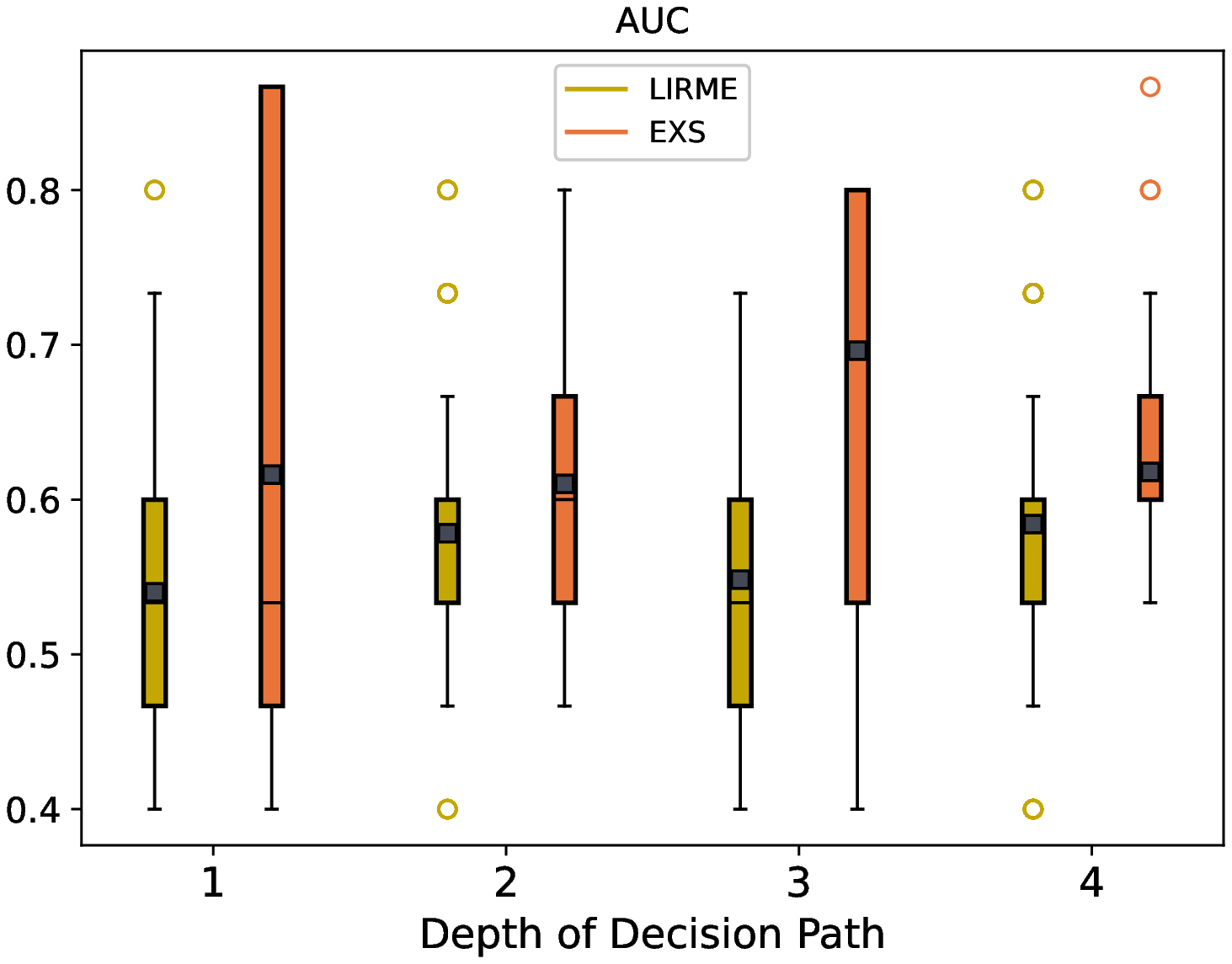}
\endminipage\hfill
\minipage{0.32\textwidth}%
  \includegraphics[width=\linewidth]{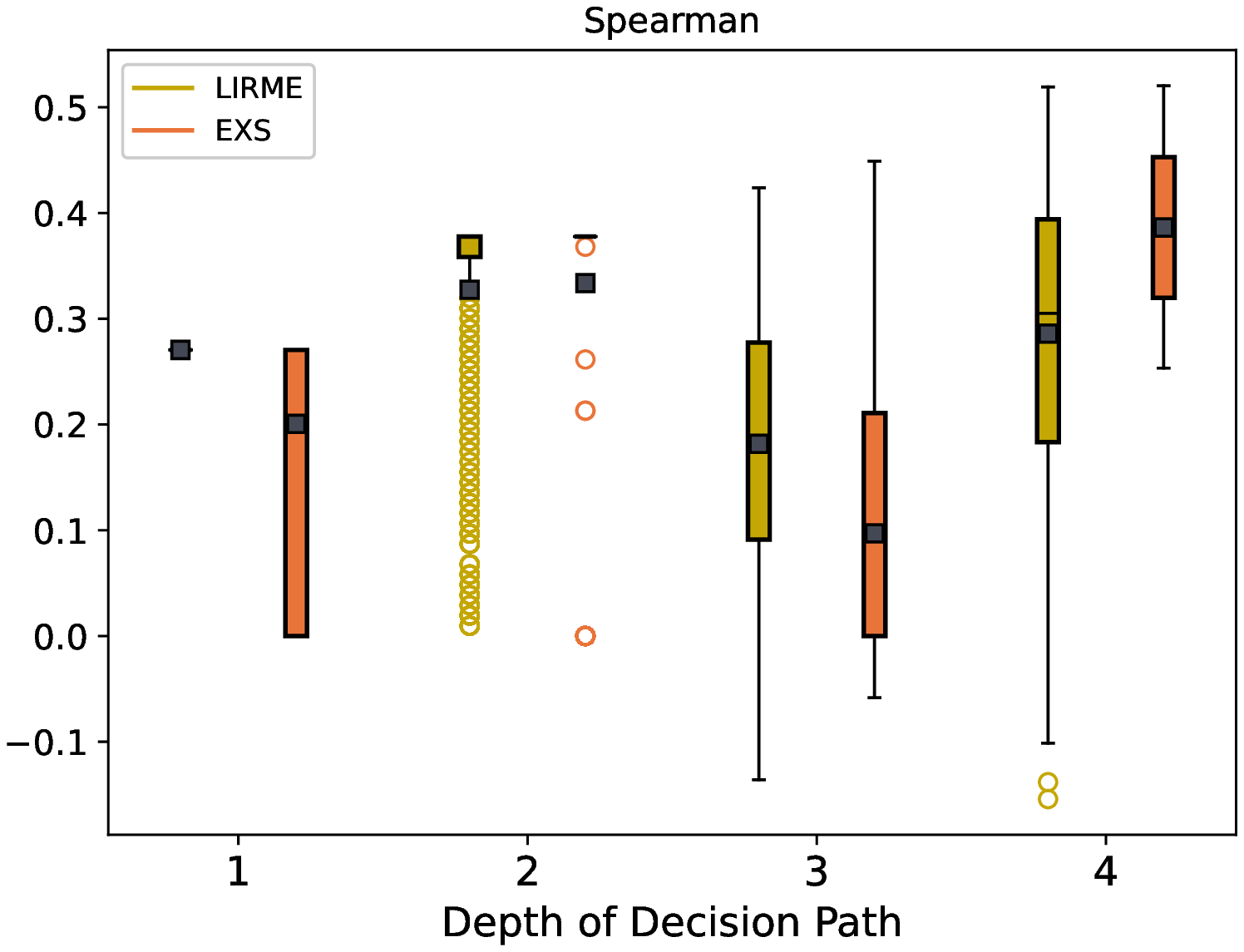}
\endminipage
\caption{Explanation accuracy based on impurity ground truth with respect to the depth of the decision path when explaining the decision tree model. The dark rectangle represents the median in each boxplot.}
 \label{fig:boxplot_path_dt_ti}
\end{figure}

 \begin{figure}
\centering
\begin{subfigure}[b]{.45\linewidth}
\includegraphics[width=\linewidth]{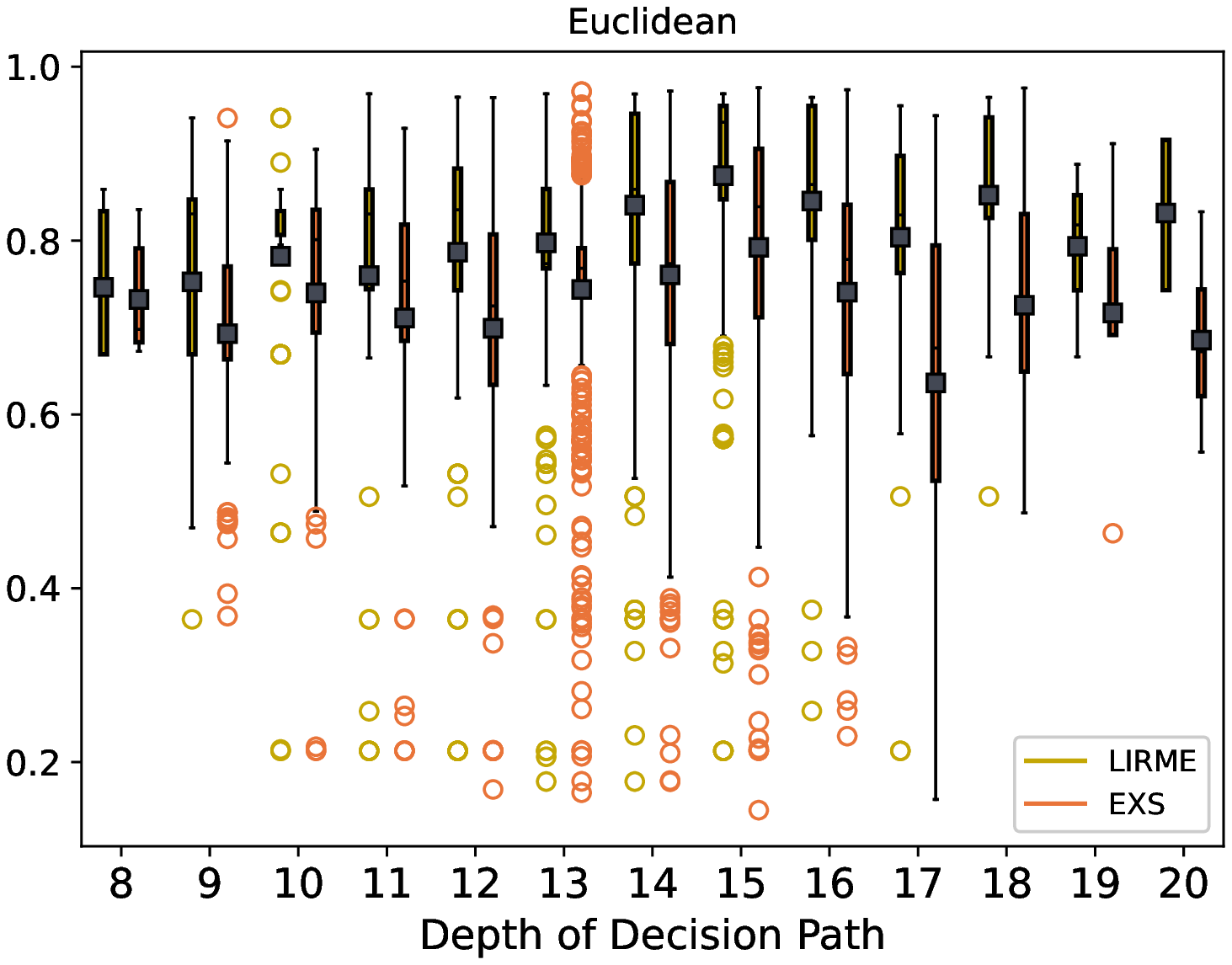}
\end{subfigure}

\begin{subfigure}[b]{.45\linewidth}
\includegraphics[width=\linewidth]{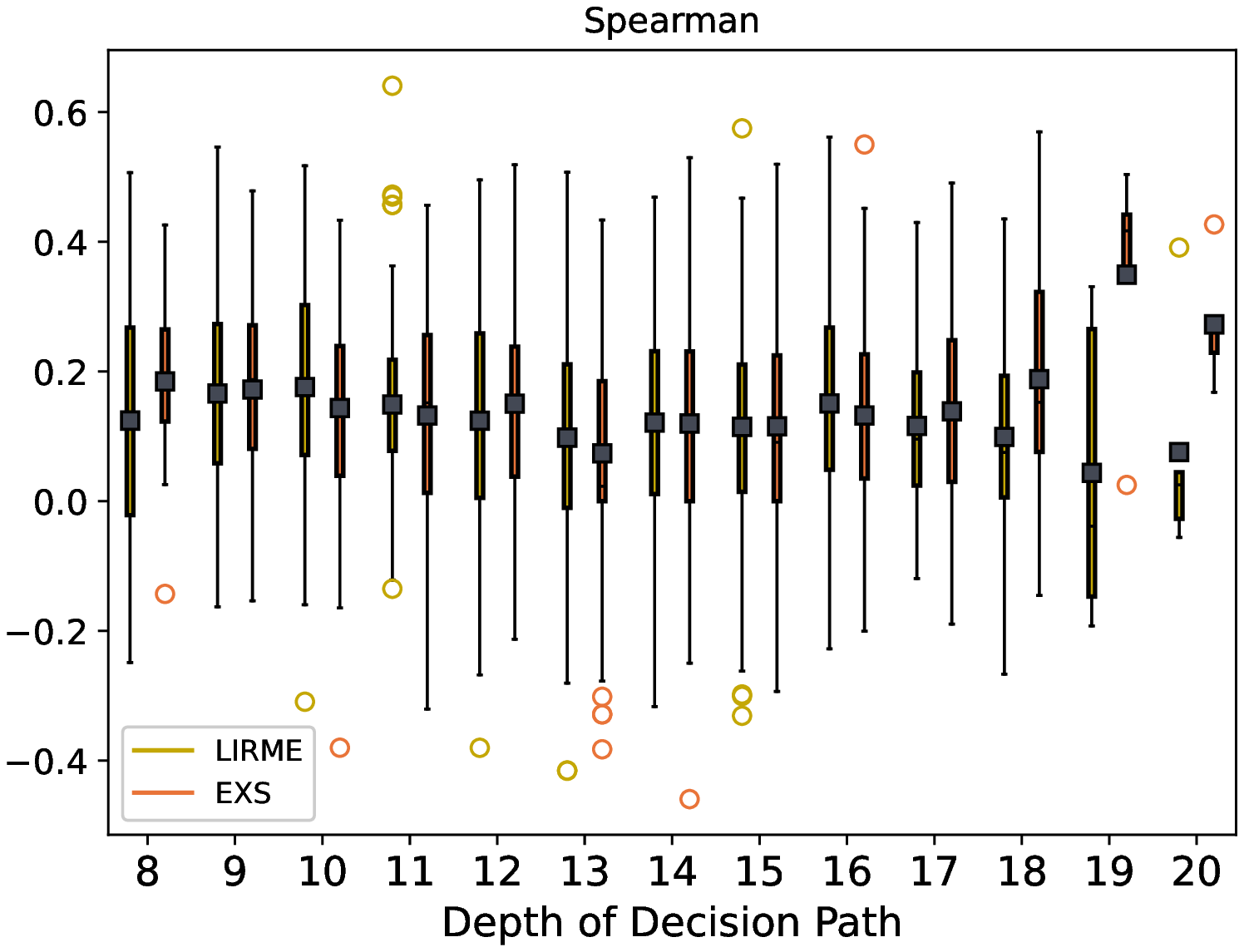}
\end{subfigure}
\begin{subfigure}[b]{.45\linewidth}
\includegraphics[width=\linewidth]{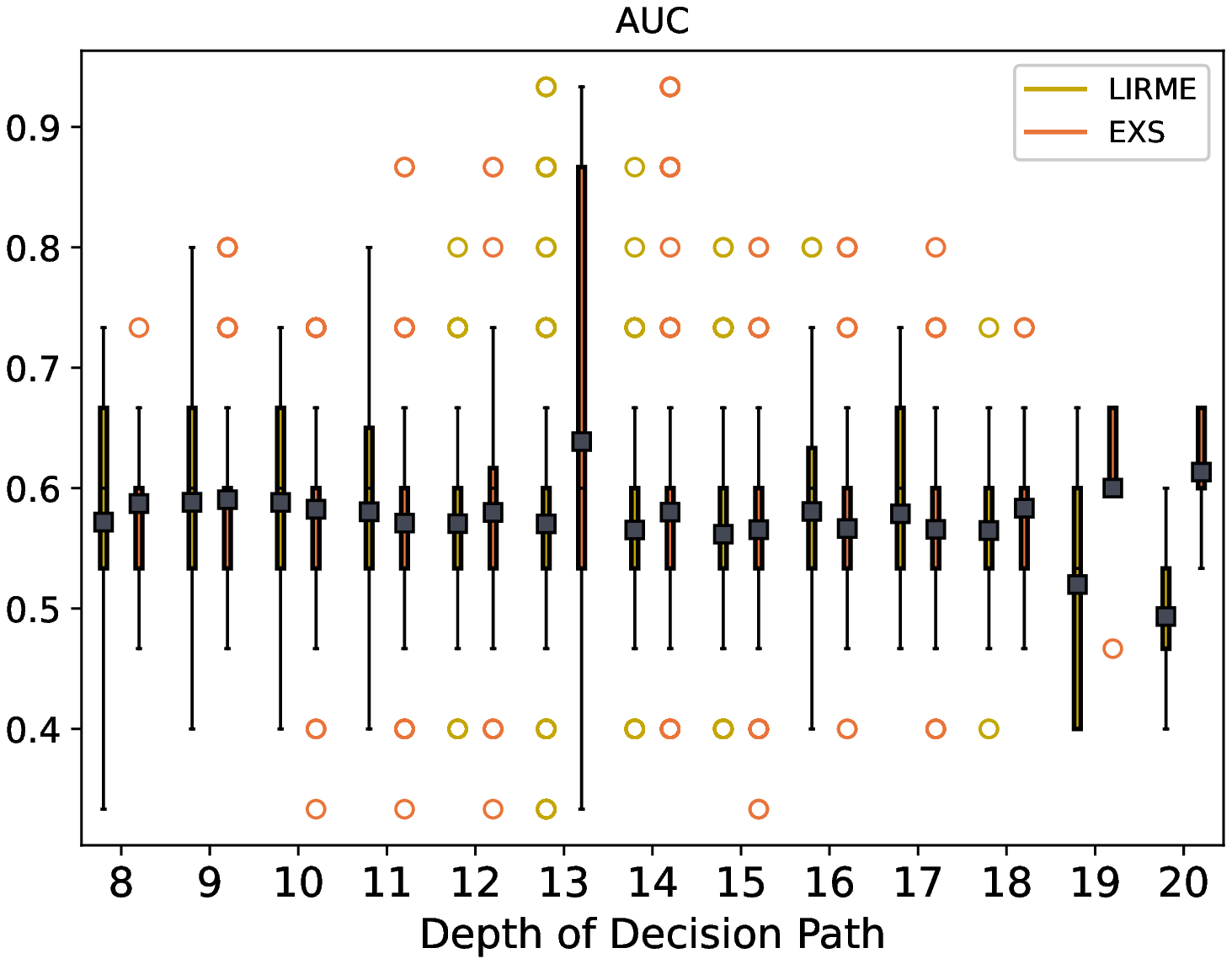}
\end{subfigure}
\caption{Explanation accuracy based on impurity ground truth with respect to the depth of the decision path when explaining the LambdaMART model. The dark rectangle represents the median in each boxplot.}
 \label{fig:boxplot_path_lmart_ti}
\end{figure}

\subsection{Experiment Details}
\subsection{Decision Tree model}
The decision tree model is selected by cross validation. The hyper-parameter optimization is performed in 100 random trials where the minimum sample split parameter had the range of 0.1 to 1. The minimum sample leaf had the range of values between 0.1 and 0.5. The maximum depth had the range of values between 1 and 20. The best estimator achieved the mean squared error of 0.241. The chosen model had min sample split and minimum sample leaf equal to 0.1 and the depth of 4.

\subsection{Lambda MART}
The decision tree model is selected by cross validation. The hyper-parameter optimization is performed in 100 random trials where the minimum sample split parameter had the range of 0.1 to 1. The minimum sample leaf had the range of values between 0.1 and 0.5. The maximum depth had the range of values between 5 and 40. The best estimator achieved the NDCG of  0.0667. The chosen model had min sample split and minimum sample leaf equal to 0.15 and the depth of 20.

\end{document}